%% file: paper.tex
\pdfoutput=1
\documentclass[11pt,letterpaper]{article}
\usepackage{emnlp2017}
\usepackage{amsmath}
\usepackage{amssymb}
\usepackage{times}
\usepackage{latexsym}
\usepackage{graphicx}
\usepackage{booktabs}
\usepackage{multirow}
\usepackage{esvect}

\newcommand{\tb}[1] {\textbf{#1}}
\newcommand{\ti}[1] {\textit{#1}}
\newcommand{\tm}[1] {\texttt{#1}}
\newcommand{\tu}[1] {\underline{#1}}

\newcommand{\MC}[3]{\multicolumn{#1}{#2}{#3}}
\newcommand{\MR}[3]{\multirow{#1}{#2}{#3}}

\newcommand{\ra}{$\rightarrow$}

\newcommand{\bvv}[1]{\reflectbox{\ensuremath{\vv{\reflectbox{\ensuremath{#1}}}}}}

\emnlpfinalcopy

\def\emnlppaperid{70}

\title{LIUM-CVC Submissions for WMT17 Multimodal Translation Task}

\author{Ozan Caglayan$^\dagger$, Walid Aransa, Adrien Bardet, Mercedes Garc\'ia-Mart\'inez, \\
         \bf Fethi Bougares, Lo\"ic Barrault \\
         LIUM, University of Le Mans \\
         $^\dagger$\tm{ozancag@gmail.com} \\
         \tm{FirstName.LastName@univ-lemans.fr} \\
          \\
         \bf Marc Masana, Luis Herranz and Joost van de Weijer\\
		CVC, Universitat Autonoma de Barcelona\\
    {\tt \{joost,mmasana,lherranz\}@cvc.uab.es}}

\begin{document}
\maketitle

\begin{abstract}
  This paper describes the monomodal and multimodal Neural Machine Translation systems
  developed by LIUM and CVC for WMT17 Shared Task on Multimodal Translation.
  We mainly explored two multimodal architectures where either global visual features
  or convolutional feature maps are integrated in order to benefit from visual context.
  Our final systems ranked first for both En$\rightarrow$De and En$\rightarrow$Fr
  language pairs according to the automatic evaluation metrics METEOR and BLEU.
\end{abstract}

\section{Introduction}
\input{introduction}
\section{Data}
\label{sec:data}
We use the Multi30k \cite{Elliott2016} dataset provided by the organizers which contains 29000, 1014 and 1000
English$\rightarrow$\{German,French\} image-caption pairs respectively for training, validation and
Test2016 (the official evaluation set of WMT16 campaign) set.
Following task rules we normalized punctuations, applied tokenization and lowercasing.
A Byte Pair Encoding (BPE) model \cite{sennrich2015neural} with 10K merge operations is learned for
each language pair resulting in 5234$\rightarrow$7052 tokens for English$\rightarrow$German
and 5945$\rightarrow$6547 tokens for English$\rightarrow$French respectively.

We report results on Flickr Test2017 set containing 1000 image-caption pairs and
the optional MSCOCO test set of 461 image-caption pairs which is considered as an \ti{out-of-domain}
set with ambiguous verbs.

\paragraph{Image Features}
We experimented with several types of visual representation using deep features
extracted from convolutional neural networks (CNN) trained on large visual datasets.
Following the current state-of-the-art in visual representation, we used a network
with the ResNet-50 architecture \cite{he2016resnet} trained on the ImageNet dataset
\cite{ILSVRC15} to extract two types of features: the 2048-dimensional features
from the \ti{pool5} layer and the 14x14x1024 features from the \ti{res4f\_relu} layer.
Note that the former is a global feature while the latter is a feature map with
roughly localized spatial information.
\section{Architecture}
\label{sec:architecture}
\input{architecture}
\section{Training}
\label{sec:training}
We use ADAM \cite{kingma2014adam} with a learning rate of $4e\mathrm{-}4$ and a batch size of 32.
All weights are initialized using Xavier method \cite{glorotxavier} and the total gradient norm is clipped
to 5 \cite{pascanu2013difficulty}.
Dropout \cite{srivastava2014dropout} is enabled after source embeddings $X$,
source annotations $\mathbf{S}$ and pre-softmax activations $o_t$ with dropout
probabilities of $(0.3, 0.5, 0.5)$ respectively. ($(0.2, 0.4, 0.4)$ for En$\rightarrow$Fr.)
An L$_2$ regularization term with a factor of $1e\mathrm{-}5$ is also applied
to avoid overfitting unless otherwise stated.
Finally, we set E=128 and R=256 (Section~\ref{sec:architecture}) respectively for embedding
and GRU dimensions.


\begin{table*}[!htbp]
\centering
\renewcommand\arraystretch{1.1}
\resizebox{.98\textwidth}{!}{%
\begin{tabular}{lcllll}
\toprule
\MR{2}{*}{\rm{En$\rightarrow$De Flickr}} & \MR{2}{*}{\# Params} & \MC{2}{c}{\rm{Test2016} ($\mu\pm\sigma/$Ensemble)}
                                         & \MC{2}{c}{\rm{Test2017} ($\mu\pm\sigma/$Ensemble)} \\
& & \MC{1}{c}{\rm{BLEU}} & \MC{1}{c}{\rm{METEOR}} & \MC{1}{c}{\rm{BLEU}} & \MC{1}{c}{\rm{METEOR}} \\ \midrule
\newcite{caglayan2016does} &  62.0M & 29.2                  & 48.5                            &                       &                         \\
\newcite{huang2016attention} & -     & 36.5                  & 54.1                            &                       &                         \\
\newcite{calixto2017doubly}  & 213M  & 36.5                  & 55.0                         &                       &                          \\
\newcite{calixto2017incorporating} & - & 37.3                & 55.1                         &                       &                         \\
\newcite{elliott2017imagination} & - & 36.8                  & 55.8                         &                       &                         \\ \midrule \midrule
Baseline NMT              & 4.6M  & 38.1 $\pm$ 0.8 / 40.7 & 57.3 $\pm$ 0.5 / 59.2           & 30.8 $\pm$ 1.0 / 33.2 & 51.6 $\pm$ 0.5 / 53.8   \\
(D1) fusion-conv          & 6.0M & 37.0 $\pm$ 0.8 / 39.9 & 57.0 $\pm$ 0.3 / 59.1           & 29.8 $\pm$ 0.9 / 32.7 & 51.2 $\pm$ 0.3 / 53.4             \\
(D2) dec-init-ctx-trg-mul & 6.3M & 38.0 $\pm$ 0.9 / 40.2 & 57.3 $\pm$ 0.3 / 59.3           & 30.9 $\pm$ 1.0 / 33.2 & 51.4 $\pm$ 0.3 / 53.7             \\
(D3) dec-init             & 5.0M & 38.8 $\pm$ 0.5 / 41.2 & 57.5 $\pm$ 0.2 / 59.4           & 31.2 $\pm$ 0.7 / 33.4 & 51.3 $\pm$ 0.3 / 53.2             \\
(D4) encdec-init          & 5.0M & 38.2 $\pm$ 0.7 / 40.6 & 57.6 $\pm$ 0.3 / 59.5           & 31.4 $\pm$ 0.4 / 33.5 & \tu{51.9} $\pm$ 0.4 / 53.7 \\
(D5) ctx-mul              & 4.6M & 38.4 $\pm$ 0.3 / 40.4 & \tu{57.8} $\pm$ 0.5 / 59.6      & 31.1 $\pm$ 0.7 / 33.5 & \tu{51.9} $\pm$ 0.2 / 53.8 \\
\tb{(D6) trg-mul}         & 4.7M & 37.8 $\pm$ 0.9 / 41.0 & \tu{57.7} $\pm$ 0.5 / \tb{60.4} & 30.7 $\pm$ 1.0 / 33.4 & \tu{52.2} $\pm$ 0.4 / \tb{54.0} \\
\bottomrule
\end{tabular}}
\caption{Flickr En$\rightarrow$De results: underlined METEOR scores are from systems significantly different ($p$-value $\leq0.05$) than the baseline
  using the approximate randomization test of \ti{multeval} for 5 runs. \tb{(D6)} is the official submission of LIUM-CVC.}
\label{tbl:ende_flickr}
\end{table*}
All models are implemented and trained with the \ti{nmtpy}
framework\footnote{\url{https://github.com/lium-lst/nmtpy}} \cite{nmtpy}
using Theano v0.9 \cite{theano}.
Each experiment is repeated with 5 different seeds to mitigate the variance
of BLEU \cite{Papineni:2002:acl} and METEOR \cite{Lavie:2007:acl} and to benefit from ensembling.
The training is early stopped if validation set METEOR does not improve for
10 validations performed per 1000 updates. A beam-search with a beam size of 12 is used for translation decoding.

\section{Results}
\label{sec:results}
All results are computed using \ti{multeval} \cite{clark2011better} with tokenized
sentences.

\input{results_en_de}

\input{results_en_fr}
\section{Conclusion}
\label{sec:conclusion}

We have presented the LIUM-CVC systems for English to German and English to French Multimodal Machine Translation evaluation campaign.
Our systems were ranked first for both tasks in terms of automatic metrics.
Using the \ti{pool5} global visual features resulted in a better performance compared to multimodal attention architecture
which makes use of convolutional features.
This might be explained by the fact that the attention mechanism over spatial feature vectors cannot capture useful information
from the extracted features maps.
Another explanation for this is that source sentences contain most necessary information to produce the translation
and the visual content is only useful to disambiguate a few specific cases.
We also believe that reducing the number of parameters aggressively to around 5M
allowed us to avoid overfitting leading to better scores in overall.

\section*{Acknowledgments}
This work was supported by the French National Research Agency (ANR) through the CHIST-ERA M2CR project\footnote{\url{http://m2cr.univ-lemans.fr}}, under the contract number ANR-15-CHR2-0006-01 and by MINECO through APCIN 2015 under the contract number PCIN-2015-251.

\bibliography{paper}
\bibliographystyle{emnlp_natbib}

\end{document}

%% file: introduction.tex
With the recent advances in deep learning, purely neural approaches to machine translation,
such as Neural Machine Translation (NMT),
\cite{Sutskever2014,Bahdanau2014} have received a lot of attention because of their
competitive performance \cite{nmteval}. Another reason for the popularity of NMT
is its flexible nature allowing researchers to fuse
auxiliary information sources in order to design sophisticated networks like
multi-task, multi-way and multi-lingual systems to name a few \cite{luong2015multi,gmlnmt,firat2016multi}.

Multimodal Machine Translation (MMT) aims to achieve better
translation performance by visually grounding the textual representations.
Recently, a new shared task on Multimodal Machine Translation
and Crosslingual Image Captioning (CIC) was proposed along with WMT16 \cite{specia2016shared}.
In this paper, we present MMT systems jointly designed by LIUM and CVC for the second
edition of this task within WMT17.

Last year we proposed a multimodal attention mechanism where two different
attention distributions were estimated over textual and image representations
using \ti{shared} transformations \cite{caglayan2016does}. More specifically,
convolutional feature maps extracted from a ResNet-50 CNN \cite{he2016resnet}
pre-trained on the ImageNet classification task \cite{ILSVRC15} were used to represent
visual information. Although our submission
ranked first among multimodal systems for CIC task, it was not able to improve
over purely textual NMT baselines in neither tasks \cite{specia2016shared}.
The winning submission for MMT \cite{caglayan2016does} was a phrase-based MT system rescored using
a language model enriched with FC$_7$ global visual features extracted from a pre-trained
VGG-19 CNN \cite{simonyan2014very}.

State-of-the-art results were obtained after WMT16 by using a \ti{separate} attention mechanism for
different modalities in the context of CIC \cite{caglayan2016multiatt} and
MMT \cite{calixto2017doubly}.
Besides experimenting with multimodal attention,
\newcite{calixto2017doubly} and \newcite{libovicky2017attention} also proposed
a gating extension inspired from \newcite{xu2015show} which is believed to
allow the decoder to learn \ti{when to attend}
to a particular modality although \newcite{libovicky2017attention}
report no improvement over baseline NMT.

There have also been attempts to benefit from different types of visual information
instead of relying on features extracted from a CNN pre-trained on ImageNet.
One such study from \newcite{huang2016attention} extended the sequence of
source embeddings consumed by the RNN with several regional features extracted
from a region-proposal network \cite{rpn}. The architecture thus predicts
a single attention distribution over a sequence of mixed-modality representations
leading to significant improvement over their NMT baseline.

More recently, a radically different multi-task architecture
called \ti{Imagination} \cite{elliott2017imagination}
is proposed to learn visually grounded representations by sharing an encoder
between two tasks: a classical encoder-decoder NMT and a visual feature reconstruction
using as input the source sentence representation.

This year, we experiment\footnote{A detailed tutorial for reproducing the results of this paper is provided at
\url{https://github.com/lium-lst/wmt17-mmt}.} with both convolutional and global visual vectors provided by the organizers
to better exploit multimodality (Section~\ref{sec:architecture}). Data preprocessing for both English$\rightarrow$\{German,French\}
and training hyper-parameters are detailed respectively in Section~\ref{sec:data} and
Section~\ref{sec:training}. The results based on automatic evaluation metrics are reported in Section~\ref{sec:results}.
The paper ends with a discussion in Section~\ref{sec:conclusion}.

%% file: architecture.tex

Our baseline NMT is an attentive encoder-decoder \cite{Bahdanau2014}
variant with a Conditional GRU (CGRU) \cite{cgru} decoder.

Let us denote source and target sequences $X$ and $Y$ with respective lengths $M$ and $N$
as follows where $x_i$ and $y_j$ are embeddings of dimension E:
\begin{align*}
  X &= (x_{1},\ldots, x_{M})\\
  Y &= (y_{1},\ldots, y_{N})
\end{align*}

\paragraph{Encoder}
Two GRU \cite{Chung2014} encoders with R hidden units each, process the source
sequence $X$ in forward and backward directions. Their hidden states
are concatenated to form a set of \ti{source annotations} $\mathbf{S}$ where each element $s_{i}$
is a vector of dimension $C=2\times R$:
\begin{equation*}
\renewcommand\arraystretch{1.5}
\mathbf{S} = \begin{bmatrix}
\mathrm{GRU_{Forw}}(\vv{X}) \\
\mathrm{GRU_{Back}}(\bvv{X})
  \end{bmatrix} \in\mathbb{R}^{M\times C}
\end{equation*}

Both encoders are equipped with layer normalization \cite{ba2016layer} where each
hidden unit adaptively normalizes its incoming activations with a learnable gain and bias.

\paragraph{Decoder}
A decoder block namely CGRU (two stacked GRUs where the hidden state of the first
GRU is used for attention computation) is used to estimate a probability distribution
over target tokens at each decoding step $t$.

The hidden state $h_0$ of the CGRU is initialized using a non-linear transformation of the
average source annotation:
\begin{equation}
  \label{eq:cgru_init}
  h_0 = \tanh\left(\mathbf{W_{init}} \cdot \frac{1}{M}{\sum_i^M s_i} \right), s_i \in \mathbf{S}
\end{equation}

\paragraph{Attention}
At each decoding timestep $t$,
an unnormalized attention score $g_i$ is computed for each source annotation $s_i$
using the first GRU's hidden state $h_t$ and $s_i$ itself:\\
($\mathrm{W_a}\in\mathbb{R}^{C}$, $\mathbf{W_s}\in\mathbb{R}^{C\times C}$ and $\mathbf{W_h}\in\mathbb{R}^{C\times R}$)
\begin{align}
  \label{eq:attscore}
  g_i = \mathrm{W_{a}^T}\tanh\left(\mathbf{W_s} s_i + b_s +\mathbf{W_h}h_t\right) + b_{a}
\end{align}
The context vector $c_t$ is a weighted sum of $s_i$ and its respective
attention probability $\alpha_i$ obtained using a softmax operation over all
the unnormalized scores:
\begin{align*}
  \alpha_i &= \text{softmax}\left([g_1,g_2,\ldots,g_M]\right)_i \\
  c_t &= \sum_i^M {\alpha}_i s_i
\end{align*}
The final hidden state $\widetilde{h}_t$ is computed by the second GRU
using the context vector $c_t$ and the hidden state of the first GRU $h_t$.

\paragraph{Output}
The probability distribution over the target tokens is conditioned on
the previous token embedding $y_{t-1}$, the hidden state of the decoder $\widetilde{h}_t$ and
the context vector $c_t$, the latter two transformed with $\mathbf{W_{dec}}$ and $\mathbf{W_{ctx}}$
respectively:
\begin{gather*}
  o_t = \tanh (y_{t-1} + \mathbf{W_{dec}}\widetilde{h}_t + \mathbf{W_{ctx}}c_t)\\
  P(y_t| y_{t-1}, \widetilde{h}_t, c_t) = \text{softmax}(\mathbf{W_{o}} o_t)
\end{gather*}

\subsection{Multimodal NMT}
\subsubsection{Convolutional Features}
The \tb{fusion-conv} architecture extends the CGRU decoder to a multimodal
decoder \cite{caglayan2016multiatt} where convolutional feature maps of 14x14x1024 are regarded as
196 spatial annotations $s_j'$ of 1024-dimension each.
For each spatial annotation, an unnormalized attention score $g_j'$ is computed
(Equation~\ref{eq:attscore}) except that the weights and biases are
specific to the visual modality and thus \ti{not shared} with the textual attention:
\begin{align*}
  g_j' = \mathrm{W_{a}'^{\ T}}\tanh\left(\mathbf{W_s}' s_j' + b_s' +\mathbf{W_h}'h_t\right) + b_{a}'
\end{align*}
The visual context vector $v_t$ is computed as a weighted sum of the
spatial annotations $s_j'$ and their respective attention probabilities $\beta_j$:
\begin{align*}
  \beta_j &= \text{softmax}\left([g_1',g_2',\ldots,g_{196}']\right)_j \\
  v_t &= \sum_j^{196} {\beta}_j s_j'
\end{align*}
The output of the network is now conditioned on a \ti{multimodal} context vector
which is the concatenation of the original context vector $c_t$ and the
newly computed visual context vector $v_t$.

\subsubsection{Global pool5 Features}
In this section, we present 5 architectures guided with
global 2048-dimensional visual representation $V$ in different ways.
In contrast to the baseline NMT, the decoder's hidden state $h_0$ is initialized
with an all-zero vector unless otherwise specified.


\paragraph{dec-init} initializes the decoder with $V$ by replacing Equation~\ref{eq:cgru_init}
with the following:
\begin{equation*}
  h_0 = \tanh\left(\mathbf{W_{img}} \cdot V\right)
\end{equation*}
\citep{calixto2017incorporating} previously explored a similar configuration (IMG$_\mathrm{D}$)
where the decoder is initialized with the sum of global visual features
extracted from FC7 layer of a pre-trained VGG-19 CNN and the last source annotation.

\paragraph{encdec-init} initializes the bi-directional encoder and the decoder with $V$
where $e_0$ represents the initial state of encoder
(Note that in the baseline NMT, $e_0$ is an all-zero vector) :
\begin{equation*}
  e_0 = h_0 = \tanh\left(\mathbf{W_{img}} \cdot V\right)
\end{equation*}

\paragraph{ctx-mul} modulates each source annotation $s_i$ with $V$
using element-wise multiplication:
\begin{equation*}
  s_i = s_i \odot \tanh\left(\mathbf{W_{img}} \cdot V\right)
\end{equation*}

\paragraph{trg-mul} modulates each target embedding $y_j$ with $V$
using element-wise multiplication:
\begin{equation*}
  y_j = y_j \odot \tanh\left(\mathbf{W_{img}} \cdot V\right)
\end{equation*}

\paragraph{dec-init-ctx-trg-mul} combines the latter two architectures
with \ti{dec-init} and uses separate transformation layers for each of them:
\begin{align*}
  h_0 &= \tanh\left(\mathbf{W_{img}} \cdot V\right) \\
  s_i &= s_i \odot \tanh\left(\mathbf{W_{img}'} \cdot V\right) \\
  y_j &= y_j \odot \tanh\left(\mathbf{W_{img}''} \cdot V\right)
\end{align*}

%% file: results_en_de.tex
\subsection{En$\rightarrow$De}

Table~\ref{tbl:ende_flickr} summarizes BLEU and METEOR
scores obtained by our systems. It should be noted that since we
trained each system with 5 different seeds, we report results obtained
by ensembling 5 runs as well as the mean/deviation over these 5 runs.
The final system to be submitted is selected based on ensemble Test2016 METEOR.

First of all, multimodal systems which use global \ti{pool5} features \ti{generally}
obtain comparable scores which are better than the baseline NMT in contrast
to \tb{fusion-conv} which fails to improve over it.
Our submitted system (D6) achieves an ensembling score of
60.4 METEOR which is 1.2 better than NMT. Although the improvements
are smaller, (D6) is still the best system on Test2017 in terms of
ensembling/mean METEOR scores. One interesting point to be stressed at this
level is that in terms of mean BLEU, (D6) performs worse than baseline on
both test sets. Similarly, (D3) which has the best BLEU on Test2016,
is the worst system on Test2017 according to METEOR. This is clearly
a discrepancy between these metrics where an improvement in one does not necessarily
yield an improvement in the other.

\begin{table}[htb]
\centering
\renewcommand\arraystretch{1.1}
\resizebox{\columnwidth}{!}{%
\begin{tabular}{lll}
\toprule
\MR{2}{*}{\rm{En$\rightarrow$De}} & \MC{2}{c}{MSCOCO ($\mu\pm\sigma/$Ensemble)}  \\
                                  & \MC{1}{c}{\rm{BLEU}} & \MC{1}{c}{\rm{METEOR}} \\ \midrule
     Baseline NMT          & 26.4 $\pm$ 0.2 / 28.7    & 46.8 $\pm$ 0.7 / 48.9             \\ \midrule
(D1) fusion-conv           & 25.1 $\pm$ 0.7 / 28.0    & 46.0 $\pm$ 0.6 / 48.0             \\
(D2) dec-init-ctx-trg-mul  & 26.3 $\pm$ 0.9 / 28.8    & 46.5 $\pm$ 0.4 / 48.5             \\
(D3) dec-init              & 26.8 $\pm$ 0.5 / 28.8    & 46.5 $\pm$ 0.6 / 48.4             \\
(D4) encdec-init           & 27.1 $\pm$ 0.9 / 29.4    & 47.2 $\pm$ 0.6 / \tb{49.2}    \\
(D5) ctx-mul               & 27.0 $\pm$ 0.7 / 29.3    & 47.1 $\pm$ 0.7 / 48.7             \\
(D6) \tb{trg-mul}          & 26.4 $\pm$ 0.9 / 28.5    & \tu{47.4} $\pm$ 0.3 / 48.8 \\
\bottomrule
\end{tabular}}
\caption{MSCOCO En$\rightarrow$De results: the best Flickr system \tb{trg-mul} (Table~\ref{tbl:ende_flickr}) has been used for this submission as well.}
\label{tbl:ende_coco}
\end{table}

\begin{table*}[htbp!]
\begin{center}
\renewcommand\arraystretch{1.1}
\resizebox{.85\textwidth}{!}{%
\begin{tabular}{lllll}
\toprule
\MR{2}{*}{\rm{En\ra Fr}} & \MC{2}{c}{Test2016 ($\mu\pm\sigma$ / Ensemble)} & \MC{2}{c}{Test2017 ($\mu\pm\sigma$ / Ensemble)}\\
 & \MC{1}{c}{BLEU}  & \MC{1}{c}{METEOR} & \MC{1}{c}{BLEU} & \MC{1}{c}{METEOR} \\
\midrule
 Baseline NMT & 52.5 $\pm$ 0.3 / 54.3 & 69.6 $\pm$ 0.1 / 71.3 			& 50.4 $\pm$ 0.9 / 53.0 	& 67.5 $\pm$ 0.7 / 69.8 \\

(F1) NMT + nol2reg & 52.6 $\pm$ 0.8 / 55.3 & 69.6 $\pm$ 0.6 / 71.7 		& 50.0 $\pm$ 0.9 / 52.5 	& 67.6 $\pm$ 0.7 / 70.0 \\

  \midrule
(F2) fusion-conv & 53.5 $\pm$ 0.8 / 56.5 & 70.4 $\pm$ 0.6 / 72.8 			& 51.6 $\pm$ 0.9 / 55.5 	& 68.6 $\pm$ 0.7 / 71.7 \\

(F3) dec-init & 54.5 $\pm$ 0.8 / 56.7 & 71.2 $\pm$ 0.4 / 73.0 					& 52.7 $\pm$ 0.9 / 55.5	& 69.4 $\pm$ 0.7 / 71.9 \\

(F4) ctx-mul & 54.6 $\pm$ 0.8 / 56.7 & 71.4 $\pm$ 0.6 / 73.0 					& 52.6 $\pm$ 0.9 / 55.7 	& 69.5 $\pm$ 0.7 / 71.9 \\

(F5) trg-mul & 54.7 $\pm$ 0.8 / 56.7 & 71.3 $\pm$ 0.6  / 73.0 					& 52.7 $\pm$ 0.9 / 55.5	& 69.5 $\pm$ 0.7 / 71.7 \\

\midrule \midrule
ens-nmt-7 & \MC{1}{r}{54.6} & \MC{1}{r}{71.6}  & \MC{1}{r}{53.3}  & \MC{1}{r}{70.1} \\
 
ens-mmt-6 & \MC{1}{r}{ \tb{57.4}} & \MC{1}{r}{\tb{73.6}} & \MC{1}{r}{\tb{55.9}} & \MC{1}{r}{\tb{72.2}} \\

\bottomrule
\end{tabular}}
\end{center}
\label{tab:enfr_flickr}
\caption{Flickr En$\rightarrow$Fr results: Scores are averages over 5 runs and given with their standard deviation ($\sigma$) and the score obtained by ensembling the 5 runs.
\ti{ens-nmt-7} and \ti{ens-mmt-6} are the submitted ensembles which correspond to the combination of 7 monomodal and 6 multimodal (global pool5) systems, respectively.}
\end{table*}

For the MSCOCO set no held-out set for model selection was available.
Therefore, we submitted the system (D6) with best METEOR on Flickr Test2016.

After scoring all the available systems (Table~\ref{tbl:ende_coco}) we observe
that (D4) is the best system according to ensemble metrics. This can be explained
by the \ti{out-of-domain/ambiguous} nature of MSCOCO
where best generalization performance on Flickr is not necessarily
transferred to this set.

Overall, (D4), (D5) and (D6) are the top systems according to METEOR on
Flickr and MSCOCO test sets.

%% file: results_en_fr.tex
\subsection{En$\rightarrow$Fr}

Table~\ref{tab:enfr_flickr} shows the results of our systems on the official test set of last year (Test2016) and this year (test2017).
F1 is a variant of the baseline NMT without $L_{2}$ regularization. 
F2 is a multimodal system using convolutional feature maps as visual features while 
F3 to F5 are multimodal systems using \ti{pool5} global visual features.
We note that all multimodal systems perform better than monomodal ones.

Compared to the MMT 2016 results, we can see that the fusion-conv (F2) system
with separate attention over both modalities achieve better performance than monomodal systems. 
The results are further improved by systems F3 to F5 which use \emph{pool5} global visual features.
We conjecture that the way of integrating the global visual features into these systems
does not seem to affect the final results since they all perform equally well on both test sets.

The submitted systems are presented in the last two lines of Table~\ref{tab:enfr_flickr}.
Since we did not have all 5 runs with different seeds ready by the submission deadline, heterogeneous ensembles of different architectures and different seeds were considered.
\ti{ens-nmt-7} (contrastive monomodal submission) and \ti{ens-mmt-6} (primary multimodal submission) correspond to ensembles of 7 monomodal and 6 multimodal (pool5) systems respectively.
\ti{ens-mmt-6} benefits from the heterogeneity of the included systems resulting in a slight improvement of BLEU and METEOR.

\begin{table}[htb]
\centering
\renewcommand\arraystretch{1.1}
\resizebox{\columnwidth}{!}{%
\begin{tabular}{lll}
\toprule
\MR{2}{*}{\rm{En$\rightarrow$Fr}} & \MC{2}{c}{MSCOCO ($\mu\pm\sigma$ / ensemble)}  \\
                                         & \MC{1}{c}{\rm{BLEU}} & \MC{1}{c}{\rm{METEOR}} \\ \midrule
Baseline NMT 	          	& 41.2 $\pm$1.2 / 43.3 & 61.3 $\pm$0.9 / 63.3 \\

(F1) NMT + nol2reg 		& 40.6 $\pm$1.2 / 43.5 & 61.1 $\pm$0.9 / 63.7 \\ \midrule

(F2) fusion-conv        	& 43.2 $\pm$1.2 / 45.9 & 63.1 $\pm$0.9 / 65.6 \\ 

(F3) dec-init 		      	& 43.3 $\pm$1.2 / 46.2 & 63.4 $\pm$0.9 / 66.0 \\

(F4) ctx-mul 	          	& 43.3 $\pm$1.2 / 45.6 & 63.4 $\pm$0.9 / 65.4 \\

(F5) trg-mul            		& 43.5 $\pm$1.2 / 45.5 & 63.2 $\pm$0.9 / 65.1 \\

\midrule
ens-nmt-7		    & \MC{1}{r}{43.6} &     \MC{1}{r}{63.4}  \\

ens-mmt-6       & \MC{1}{r}{45.9} &     \MC{1}{r}{65.9}  \\

\bottomrule
\end{tabular}}
\caption{MSCOCO En$\rightarrow$Fr results: \tb{ens-mmt-6}, the best performing
ensemble on Test2016 corpus (see Table~\ref{tab:enfr_flickr}) has been used for this submission as well.}
\label{tab:enfr_coco}
\end{table}

Results on the ambiguous dataset extracted from MSCOCO are presented in Table~\ref{tab:enfr_coco}.
We can observe a slightly different behaviour compared to the results in Table~\ref{tab:enfr_flickr}. 
The systems using the convolutional features are performing equally well compared to those using \ti{pool5} features.
One should note that no specific tuning was performed for this additional task since no specific validation data was provided.